\documentclass[journal]{IEEEtran}
\usepackage{xcolor,soul,framed} 
\usepackage{pifont}
\usepackage[utf8]{inputenc}
\usepackage{booktabs}
\usepackage{tabularx}
\colorlet{shadecolor}{yellow}
\usepackage[pdftex]{graphicx}
\graphicspath{{../pdf/}{../jpeg/}}
\DeclareGraphicsExtensions{.pdf,.jpeg,.png}
\newcommand{\cmark}{\ding{51}}  
\newcommand{\xmark}{\ding{55}}  
\usepackage{amssymb}
\usepackage[cmex10]{amsmath}
\usepackage{array}
\usepackage{multirow}
\usepackage{mdwmath}
\usepackage{mdwtab}
\usepackage{eqparbox}
\usepackage{url}

\begin{document}
\bstctlcite{IEEEexample:BSTcontrol}
    \title{Meta‑Thinking in LLMs via Multi‑Agent Reinforcement Learning: A Survey}
  \author{Ahsan~Bilal,~\IEEEmembership{Student Member,~IEEE,}
      Muhammad~Ahmed~Mohsin,~\IEEEmembership{Graduate Member,~IEEE,}\\
      Muhammad~Umer,~\IEEEmembership{Graduate Member, IEEE}
      Muhammad~Awais~Khan~Bangash,~\IEEEmembership{Student Member,~IEEE,}
      and~Muhammad~Ali~Jamshed,~\IEEEmembership{Senior Member,~IEEE}

  \thanks{Ahsan Bilal is with University of Oklahoma, Norman, OK, 73072, USA (e-mail: ahsan.bilal-1@ou.edu).}
  \thanks{Muhammad Ahmed Mohsin, Muhammad Umer are with Stanford University, Stanford, CA, 94305, USA (e-mail: {muahmed, mumer}@stanford.edu).}%
  \thanks{Muhammad Awais Khan Bangash is with the School of Electrical and Computer Engineering, Oklahoma State University, Stillwater, OK, 74075 USA (e-mail: awais.bangash@okstate.edu).}
  \thanks{Muhammad Ali Jamshed is with University of Glasgow, G12 8QQ, Glasgow, UK (e-mail: muhammadali.jamshed@glasgow.ac.uk).}}


\maketitle

\begin{abstract}
This survey explores the development of meta-thinking capabilities in Large Language Models (LLMs) from a Multi-Agent Reinforcement Learning (MARL) perspective. Meta-thinking—self-reflection, assessment, and control of thinking processes—is an important next step in enhancing LLM reliability, flexibility, and performance, particularly for complex or high-stakes tasks. The survey begins by analyzing current LLM limitations, such as hallucinations and the lack of internal self-assessment mechanisms. It then talks about newer methods, including RL from human feedback (RLHF), self-distillation, and chain-of-thought prompting, and each of their limitations. The crux of the survey is to talk about how multi-agent architectures, namely supervisor-agent hierarchies, agent debates, and theory of mind frameworks, can emulate human-like introspective behavior and enhance LLM robustness. By exploring reward mechanisms, self-play, and continuous learning methods in MARL, this survey gives a comprehensive roadmap to building introspective, adaptive, and trustworthy LLMs. Evaluation metrics, datasets, and future research avenues, including neuroscience-inspired architectures and hybrid symbolic reasoning, are also discussed.
\end{abstract}


%
\IEEEpeerreviewmaketitle


\section{Introduction}
\begin{figure}[ht]
  \centering
  \includegraphics[width=0.75\linewidth]{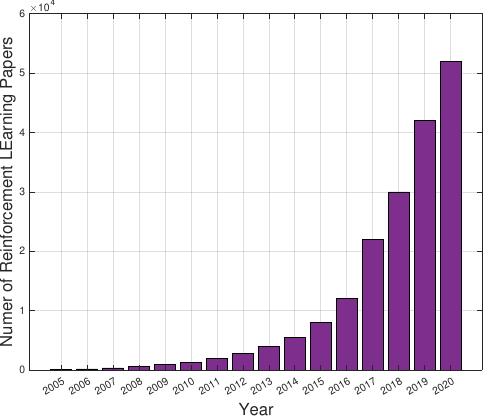}
  \caption{Annual number of RL publications in AI conferences (2019–2024).}
  \label{fig:rl_trend}
\end{figure}

\IEEEPARstart{T}{the} cognitive abilities, such as intelligence and creativity, have played a fundamental role in human discoveries and inventions. Understanding the relationship between these two cognitive abilities is important not only for the advancement of psychological theories but also for the improvement of educational practices \cite{karwowski2016creativity}. However, researchers still hold different views on how intelligence and creativity interact, often leading to conflicting findings. A key question in this discourse is how intelligence enables structured problem-solving, while creativity fosters novel solutions that are essential for human cognition and artificial intelligence systems. For example, in creative writing tasks, intelligence helps in structuring narratives and developing characters, whereas creativity drives originality and emotional depth. Similarly, in problem-solving tasks, intelligence aids in analyzing constraints, while creativity allows for flexible and unconventional approaches. Moreover, the role of internal thought processes varies with task complexity. Simpler tasks require minimal reasoning, whereas more complex tasks demand deeper cognitive engagement. This principle extends to artificial intelligence, where more sophisticated models exhibit enhanced performance in tasks requiring higher-order thinking. Based on this, researchers hypothesize that thinking LLMs will have a distinct advantage in handling sufficiently complex tasks, as they integrate structured reasoning with creative problem-solving abilities \cite{wu2024thinking}.\\
\indent One practical approach to enable this creative thinking reasoning is through text-based thought generation and leveraging the natural language capabilities of LLMs. Since, LLMs are pre-trained on vast amounts of text that include human-written reflections, they inherently encode aspects of human reasoning processes. LLMs can be prompted to engage in internal reasoning before responding, allowing for more thoughtful and structured responses \cite{bilal2025llms}. For example, the chain-of-thought (CoT) prompting technique \cite{wei2022chain} facilitates the reasoning process by guiding LLMs to articulate their reasoning steps explicitly, improving their performance on tasks that require logical and mathematical reasoning. However, the benefits of CoT prompting appear to be domain-specific. A meta-analysis by \cite{sprague2024cot} found that while CoT significantly enhances LLM performance in structured reasoning tasks, it provides little to no advantage in areas that do not involve mathematical or logical processes. \\
\indent LLMs such as GPT-3, LLaMA, and PaLM, have become transformative tools across various domains, such as natural language processing (NLP), healthcare, education, software development, and scientific research, demonstrating remarkable proficiency in diverse tasks, such as text generation, language translation, summarization, sentiment analysis, code generation, medical diagnosis assistance, and automated tutoring \cite{mohsin2025retrieval}. Despite their success, LLMs continue to face the challenge of "hallucination," where they generate content that is inaccurate or not based on factual information \cite{huang2025survey}. The issue is particularly critical in high-stakes applications, such as clinical and legal domains, where reliable and accurate text generation is essential. Ensuring reliability through self-assessment is important in today's rapidly evolving and high-stakes environment, where even small changes in the accuracy of LLMs can result in severe consequences \cite{gallagher2024assessing}. Addressing hallucinations in LLMs is crucial for expanding their practical applicability and fostering trust in these technologies. Hallucinations in LLMs are of three main categories \cite{zhang2023siren}:\\
\textbf{1) Input conflict: } in which the generated output contradicts the given prompt;\\ 
\textbf{2) Context conflict:} in which there are inconsistencies within the generated response; and\\
\textbf{3) Factual conflict:} in which LLM produce incorrect information, despite having access to accurate knowledge. \\
Understanding and mitigating these hallucinations is an important step to improve the robustness and reliability of LLMs in real-world scenarios. One of the key challenges is that LLMs generate responses in an autoregressive manner without an inherent mechanism for evaluating their own outputs and this leads to unchecked errors.\\
\indent Hallucinations in LLMs can be eliminated through number of approaches \cite{yin2024review}. One of the most popular methods is using \textbf{RL}, where the model is trained to give better responses based on feedback. For example, ChatGPT-3.5 uses a method known as RLHF \cite{singhal2023long}, where human assessors rank the responses of the model, training it on what sounds right. RL is effective in modern AI that enables agents to learn complex behavior by trial-and-error exploration of their environment. Its capacity to discover new policies without explicit supervision has spurred progress in robotics, game playing, and autonomous systems \cite{mnih2015human, lillicrap2015continuous}. At the same time, the scale of RL work has grown exponentially: annual papers in top-tier venues have increased by over 400\% in the last half-decade, as an indicator of a continuously accelerating interest in RL methods (see Figure \ref{fig:rl_trend}). This growing pattern emphasizes the imperative of integrating RL, especially multi-agent variants, into meta-thought models of LLMs. More recently, frameworks like DeepSeek have built on this concept further \cite{guo2025deepseek}. Instead of only using RL towards the later stages of training, they use it throughout the training process. This way, the LLM is learning from itself throughout and improving the extent to which it follows commands, thinks for itself, and stays consistent with what people expect. Another method is \textbf{contrastive learning}, as discussed in MixCL \cite{zhang2024mixcl}, where the LLM is shown both good and bad examples to help it better understand the difference between true and false information. \textbf{Knowledge-based} methods which leverage structured external sources to detect and correct hallucinations. Knowledge-based methods externally are Neural Path Hunter \cite{dziri2021neural}, Knowledge Graph Retrofitting (KGR) \cite{guan2024mitigating}, Validating Low-Confidence Generation \cite{varshney2023stitch}, and Reasoning on Graphs (RoG) \cite{luo2023reasoning}. Knowledge filling methods try to fill in the gaps in the model's internal knowledge to make its answers more accurate \cite{casper2023open}. Other approaches don't rely on external data at all and rely solely on the model itself to iteratively assess and improve its output. For example, Zero-resource and self-feedback methods like SelfCheckGPT \cite{manakul2023selfcheckgpt}, Chain-of-Verification \cite{dhuliawala2023chain}, Self-Refine \cite{zhang2024self}, and Self-Contradictory \cite{mundler2023self} help the LLM model review and improve its own response. Finally, decoding-level interventions are Knowledge-Constrained Tree Search \cite{choi2023kcts}, Inference-Time Intervention \cite{li2023inference}, and Decoding by Contrasting Layers (DoLa) \cite{chuang2309decoding}, all of which attempt to guide token generation in LLMs towards factuality. However, there are major pitfalls in each methodology: RL depends on human biased labeling and creates bias \cite{casper2023open, singhal2023long}; external knowledge retrieval is limited in terms of delay and accuracy for retrieval data \cite{casper2023open}; sampling lacks completeness of information and may miss important facts \cite{zhang2024mixcl}; and self-feedback is unreliable in zero resource environments when the model has no access to outside help \cite{madaan2023self,zhang2024self}. Most importantly, all of these methods still rely on how LLMs are generated at a basic level: by predicting one word at a time based on the words preceding it. This framework makes it possible for minor errors to build up into more significant ones, and the model does not actually "understand" truth in the same manner as humans \cite{valmeekam2023can}.
\indent To address the issue of hallucination, one of the most promising research directions is to enhance the meta-thinking (the ability to reflect on, evaluate, and regulate one’s own thought processes) capabilities of LLMs so that LLMs can analyze, control, and refine their own thought processes. By implementing self-evaluation mechanisms, LLMs can identify inconsistencies, assess the reliability of their generated content, and improve their reasoning before producing a final response \cite{dhuliawala2023chain}. Compared to human thought, in which individuals think about themselves to check for the validity of their ideas before conclusion, LLMs lack an inherent self-checking system. By embedding meta-thinking, LLMs can examine their own reasoning steps, identify potential inconsistencies, and adaptively update their responses before rendering the final output \cite{griot2025large}. To overcome the limitations imposed by the autoregressive generation process of LLMs, it is very important to integrate the meta-thinking via MARL because it empowers LLMs to collaboratively self-correct and adapt in real time, thereby \cite{huang2023large}.\\
\indent Building on these advances, MARL can offer frameworks for LLMs to develop natural language capabilities with collaborative AI behavior \cite{sun2024llm}. For example, DyLAN \cite{liu2023dynamic} supports dynamic agent selection according to task difficulty so that agents based on LLM can be dynamically configured at runtime for uses like reasoning and code composition. FAMA \cite{slumbers2023leveraging} functionally configures agents at runtime by means of fine-tuning and enables natural language based communication for uses like text-based games and simulated driving. MetaGPT \cite{hong2023metagpt} supports agents with publish and subscribe capability on task specific messages in a shared message pool for collaborative coding. In robotics, CoELA \cite{zhang2023building} integrates LLMs with modular perception, memory, planning, and execution systems to enable robots to become smarter to cooperate. SMART-LLM \cite{kannan2024smart} transforms high-level commands into robot team executable plans by breaking tasks into phases. RoCo \cite{mandi2024roco} also provides robot arms with LLM agents that interact using conversation to manage action coordination. In Co-NavGPT \cite{yu2023co}, one LLM manages navigation for multiple agents with frontier assignments optimized. The study in \cite{guo2024embodied} introduced a Criticize-Reflect architecture, assigning LLMs to perform critic and coordinator roles to enhance coordination between agents. ConsensusLLM \cite{chen2023multi} focuses on negotiation, where agents of various initial states reach agreement via natural language. In summary, as LLMs are increasingly applied in important domains, the integration of meta-thinking via MARL is a necessary evolution to ensure safety, accuracy, and adaptability in high-stakes, dynamic applications \cite{chua2025learning}.\\
\indent A meta-thinking induction framework for LLMs based on MARL was introduced in \cite{wan2025rema}. The approach introduces meta-thinking by assigning various high and low-level agents to LLMs, where exploration is encouraged during training, and explainability about the response is enhanced. Researchers evaluate LLM-based agents in a multi-agent cooperative text game with ToM inference tasks \cite{li2023theory}, which compares the performance of LLM-based agents with MARL against planning-based baselines. The results showed that LLM-based agents developed emergent collaborative behaviors and demonstrated higher-order ToM capabilities, indicating their ability to infer and respond to the thoughts and intentions of other agents. \\
\indent The main motivation behind this survey paper is that meta-reasoning and meta-thinking in LLMs have gained more and more attention. For example, \cite{yang2024buffer} introduces the meta-buffer to capture the general patterns of thoughts and change these thoughts continuously in self-reflection. Most existing works \cite{lee2022meta, hong2023metagpt, yin2024review, wu2024continual, shi2024continual, jovanovic2024towards, zheng2023learn, gao2024meta, plaat2024reasoning, sayed2024gizaml, zheng2025towards, bandyopadhyay2025thinking} focus on individual techniques like COT prompting or self-reflection, or explore multi-agent coordination in an architectural framework. But to date, there is no thorough survey that combines Meta-RL and MAS towards introspective and self-critical LLMs. This work fills that vital gap by exploring how the principles of RL, especially in multi-agent environments, can be used to make LLMs to meta-think. This paper provides the first roadmap for making dependable and responsive LLMs through meta-reasoning by poetically connecting meta-cognition, MARL frameworks, and datasets.

The main contribution of this survey is that it is the first to investigate systematically the intersection of \textbf{Meta-RL} and \textbf{meta-thinking in LLMs}. The main contributions are: 
\begin{enumerate}
    \item A general taxonomy of meta-thinking approaches for LLMs, including single-agent (e.g., Self-Distill, SelfCheckGPT, Chain-of-Verification), multi-agent (e.g., supervisor-worker hierarchies, agent-role debates, Theory-of-Mind configurations), and reinforcement-based self-improvement methods (e.g., RLHF, adaptive intrinsic reward shaping).
    \item A systematic study of MARL paradigms for meta-reasoning. This survey paper gives a thorough review of the MARL-enabled self-reflection and correction in LLMs. This includes  1) Supervisor-Agent Architectures, in which a high-level controller coordinates low-level agents for structured decomposition and synthesis (e.g., Co-NavGPT, FAMA) of tasks. 2) Adversarial Debate and Self-Critique, as inspired by DebateQA, these setups foster argumentation among agents to expose logical flaws. 3) Self-Play and Role-Play Systems,  in which environments like AutoGPT and Criticize-Reflect showcase how iterative, role-based interaction enhances strategic depth. 4) Intrinsic Reward Mechanisms, in which Meta-reward combines human feedback with curiosity-driven scoring to guide reasoning refinement.
    \item \label{sec:five_axes} A comparative survey of recent survey papers on LLM meta-reasoning, to show that none of the papers jointly address all five axes: 1) meta-cognition, 2) multi-agent design, 3) reinforcement frameworks, 4) existing datasets, and 5) emerging architectures. Our work fills this gap and highlights convergence points and research gaps. 
    \item We include the latest metrics, i.e., Error Localization Accuracy (ELA) \cite{zeng2024mr}, Depth-Wise Accuracy \cite{patel2024multi}, Meta vs. Object-Level Accuracy \cite{ferguson2025evaluating}, and Algorithm (AIA)/Malgorithm Identification Accuracy (MIA) \cite{sonkar2024malalgoqa}, and datasets and evaluation pipelines. We also propose Summary Evaluation Quality (SEQ) as a universal summarization metric derived from the METAL \cite{hada2024metal} dataset.
    \item Identification of core research challenges in MARL-enhanced meta-reasoning, including scalability (coordination bottleneck in multi-agent setups), reward hacking (model collapse in reinforcement-based training), energy efficiency (risk for autonomous adaptation), and ethical bias (in self-reinforcing feedback loops).
    \item A forward-looking roadmap proposing neuroscience-inspired architectures episodic memory, uncertainty gating, and metacognitive control modules, symbolic-MAL hybrids to combine symbolic logic with adaptive MARL agents for interpretable reasoning and safety assurance, and dynamic agent configurations to enable trustworthy, self-correcting AI.
\end{enumerate}

In this survey paper, we start by reviewing basic fundamental concepts of meta-thinking and meta-learning, followed by an in-depth discussion of current single-agent and multi-agent approaches to enable self-reflective reasoning in LLMs. Continuoing the discussion of multi-agent, we then examine various MARL strategies that support meta-reasoning, including reward mechanisms, self-play with adversarial learning. To evaluate these advances, we present key metrics, datasets, and comparative studies. Table \ref{tab:comparison} provides a comprehensive comparison of previous survey papers on meta-reasoning in LLMs, highlighting how our work uniquely integrates all five critical aspects as described in contribution no. \ref{sec:five_axes}.  Finally, we identify open research challenges and future directions that highlight the potential of neuroscience-inspired models and hybrid MARL-symbolic reasoning systems. This survey aims to provide a comprehensive roadmap for researchers seeking to develop more introspective, trustworthy, and adaptive LLMs.

\begin{table*}[h]
  \caption{Comparison of Survey Papers on Meta-Reasoning in LLMs}
  \centering
  \label{tab:comparison}
  \begin{tabular}{lccccc}
    \toprule
    \textbf{Survey Paper} & \textbf{Meta-Thinking} & \textbf{Multi-Agent} & \textbf{RL / Meta-RL} & \textbf{Eval. Benchmarks} & \textbf{Emerging Directions} \\
    \midrule
    Lee et al. (2022) \cite{lee2022meta} & \cmark & \xmark & \cmark & \cmark & \xmark \\
    MetaGPT (2023) \cite{hong2023metagpt} & \cmark & \cmark & \xmark & \xmark & \xmark \\
    Yin et al. (2024) \cite{yin2024review} & \cmark & \xmark & \cmark & \cmark & \xmark \\
    Wu et al. (2024) \cite{wu2024continual} & \cmark & \xmark & \cmark & \cmark & \cmark \\
    Shi et al. (2024) \cite{shi2024continual} & \cmark & \xmark & \cmark & \cmark & \cmark \\
    Jovanović et al. (2024) \cite{jovanovic2024towards} & \cmark & \xmark & \cmark & \cmark & \cmark \\
    Zheng et al. (2024) \cite{zheng2023learn} & \cmark & \xmark & \cmark & \cmark & \cmark \\
    Gao et al. (2024) \cite{gao2024meta} & \cmark & \xmark & \xmark & \cmark & \xmark \\
    Plaat et al. (2024) \cite{plaat2024reasoning} & \cmark & \xmark & \cmark & \cmark & \cmark \\
    Sayed et al. (2024) \cite{sayed2024gizaml} & \cmark & \xmark & \xmark & \cmark & \cmark \\
    Zheng et al. (2025) \cite{zheng2025towards} & \cmark & \xmark & \cmark & \cmark & \cmark \\
    Dibyanayan et al. (2025) \cite{bandyopadhyay2025thinking} & \cmark & \xmark & \cmark & \cmark & \cmark \\
    \textbf{Ahsan et al. (2025)} & \cmark & \cmark & \cmark & \cmark & \cmark \\
    \bottomrule
  \end{tabular}
\end{table*}\
\section{Background}
\subsection{Meta-Thinking and Meta-Learning}
Meta-thinking is the process of reflecting and analyzing one's own thinking. It is also highly associated with meta-cognition, which typically involves awareness, monitoring, and regulation of one's cognitive activities. Meta-learning, or "learning to learn," addresses the development of strategies enabling a system to easily move into new tasks by employing previous experiences. In a meta-learning configuration, an LLM may learn to change its output based on feedback from earlier interactions and, therefore, increase its performance when working on novel tasks \cite{griot2025large}. A prominent example is MetaICL (Meta-training for In-Context Learning) \cite{min2021metaicl}, where a pre-trained language model is tuned on a diverse collection of tasks in the form of input-output sequences. During this meta-training phase, the model learns to interpret and respond to new tasks presented in a similar sequence format. If afterwards it sees a task that it has never encountered before, it can employ the few examples to figure out how to proceed without needing to be retrained. This approach is good to utilize if the new task is quite different from what it was trained on.
\subsection{Meta-Thinking in LLMs:} 
Meta-reasoning in LLMs is essential to make LLMs adapt to new challenges. With the capability of self-reflection mechanisms, LLM can realize its own limitations and make changes in its reasoning strategies accordingly \cite{gao2024meta}. This is especially required for LLMs in dynamic environments where context and requirements can change rapidly. For example, when an LLM discovers that its output does not fully respond to a question posed by a user, meta-thinking processes can trigger a re-evaluation of the response or even request additional context. This kind of self-awareness in LLMs not only improves accuracy but also trustworthiness in high-stakes settings such as legal or clinical setups \cite{esteitieh2025towards}. In these environments, even a small mistake can have serious consequences, such as a wrong legal interpretation or an incorrect medical suggestion. If the LLM can spot gaps in its answers, explain its reasoning, or express uncertainty when it’s not sure, people are more likely to trust the LLM's response. For example, if an LLM response overlooks a critical symptom in a clinical use case, the meta-reasoning framework allows it to detect the oversight, reconsider the input, and revise its recommendation, refining accuracy and trust in high-stakes decision-making. Current research has explored these ideas, suggesting that more sophisticated meta-thinking frameworks in LLMs can mitigate issues like hallucinations, where models produce factually incorrect or misleading information. \\
\indent Despite significant advancements in language modeling, LLMs still face challenges in achieving true meta-thinking. Although today's models can do basic self-assessment, such as estimating confidence levels or marking uncertainty in responses. They lack the meta-cognitive ability that is required for thorough self-evaluation, error correction, and adaptive reasoning. For example, research in \cite{yin2023large} found that although LLMs can flag ambiguous queries, they struggle to engage in robust internal reasoning necessary for full meta-cognition. This gap leads to practical issues, such as the generation of hallucinated content, giving the need for further research to build LLMs that can more effectively reflect on and refine their thought processes.
\subsection{Role of MARL in Enhancing Meta-Thinking in LLMs}
\indent The integration of meta-thinking, meta-cognition, and meta-learning in LLMs is an important step to build more robust and adaptive language models. The emergence of MARL systems presents a promising direction to further improve meta-cognitive capabilities by enabling LLMs to reflect on their internal reasoning, adapt to new challenges, and collaboratively evolve their strategies \cite{li2024language} of cross-checking the responses. As research in these areas progresses, combining these ideas opens the door to building the next generation of language models that are not only smarter but also more self-aware and trustworthy. For example, framework proposed in \cite{cross2024hypothetical} allows LLMs to predict and identify the action of other agents in a multi-agent system. The framework includes a ToM component that enables the model to make assumptions and refine its assumptions about other agents' strategies. The LLM makes a guess, observes the outcome, and then adjusts its assumptions accordingly. This continous process of learning allows the LLM to improve its decision-making capabilities over time.  Through repeated experiments, the model learns to make better decisions, i.e., demonstrating a more advanced level of reasoning, similar to how humans improve through experience.
\section{Meta-Thinking Frameworks}
This section discusses three primary categories of methodologies to install meta-thinking in LLMs: (1) Single-Agent Methods, (2) Multi-Agent Architectures for Meta-Thinking, and (3) Emerging Methods in Self-Improvement. While each approach varies in design and scope, they collectively highlight an evolving research trend to enable models to reason about their reasoning processes.
\subsection{Meta-Thinking Techniques using Single-Agent Methods}
\begin{figure*}[ht]
    \centering
    \includegraphics[width=0.9\linewidth]{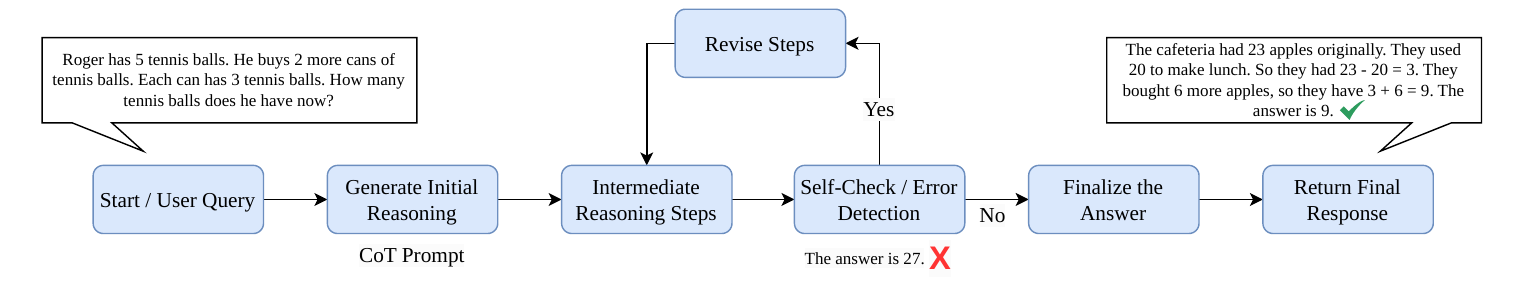}
    \caption{A COT flowchart adapted from the arithmetic examples in Wei \textit{et al.} \cite{wei2022chain}.
    The model uses explicit reasoning steps, checks for errors, and revises before returning the final result.}
    \label{fig:cot_flowchart}
\end{figure*}

Single-agent method helps to develop meta-thinking in LLMs through the methods of self-distillation, reflective prompting, and CoT reasoning. In self-distillation, an LLM generates "teacher" responses that guide a "student" version of itself towards improved performance \cite{hinton2015distilling}. Unlike supervised learning, which is based on external labeled data (e.g., human-annotated responses), self-distillation enables the model to learn from its own outputs. For example, in supervised learning, a model may be trained to respond to a question based on a dataset in which the human-annotated correct responses are given. In contrast, with self-distillation, the model responds to the question by itself, then bootstraps its future responses by learning from its previous (and potentially better) output. New forms of self-distillation aim to improve the model's own reasoning abilities \cite{gou2021knowledge}. As LLMs can detect inconsistencies or mistakes and modify future responses accordingly by repeatedly going back over their responses. \\
\indent Symmetrically, reflective prompting has also become another means of inducing meta-thinking in LLMs \cite{liu2024large}. It involves prompting an LLM to generate explicit "reflections" of its reasoning steps in the guise of a narrative of how it arrived at some conclusion. These reflections not only provide insight into the model's thought process but also a chance for the model to critique or revise earlier steps. Aside from reflection-based prompting, CoT reasoning explicitly breaks down difficult problems into intermediate steps \cite{wei2022chain} and uncovers the decision-making paths of the LLMs. We illustrate the CoT workflow with a simple arithmetic example from \cite{wei2022chain}. Figure \ref{fig:cot_flowchart} shows how an LLM walks through initial reasoning, intermediate steps, self-check, and revision before finalizing the answer. But recent studies show that LLMs struggle with self-evaluation, they often stick to their first line of thinking, even if it’s wrong, and won’t go back to question their steps unless you clearly ask them to do so \cite{chen2022can}.\\
\indent Despite all these advances, there remain limitations in self-improvement and self-assessment. If internal feedback loops are not properly controlled, single-agent methods reinforce flawed patterns of reasoning as a side effect. The model's ability to identify inconsistency or correct a poor line of argument may therefore be limited by the scope and volume of its own internally derived data \cite{andreas2019measuring}. For example, a language model trained on biased financial data can consistently overestimate market stability, reinforcing incorrect predictions without external verification. These constraints highlight the need for more robust systems, e.g., external input or multi-agent processes, to mitigate the inherent risks of the "echo chamber" phenomenon of self-enhancing methods.
\begin{figure*}[h]
    \centering
    \includegraphics[width=\linewidth]{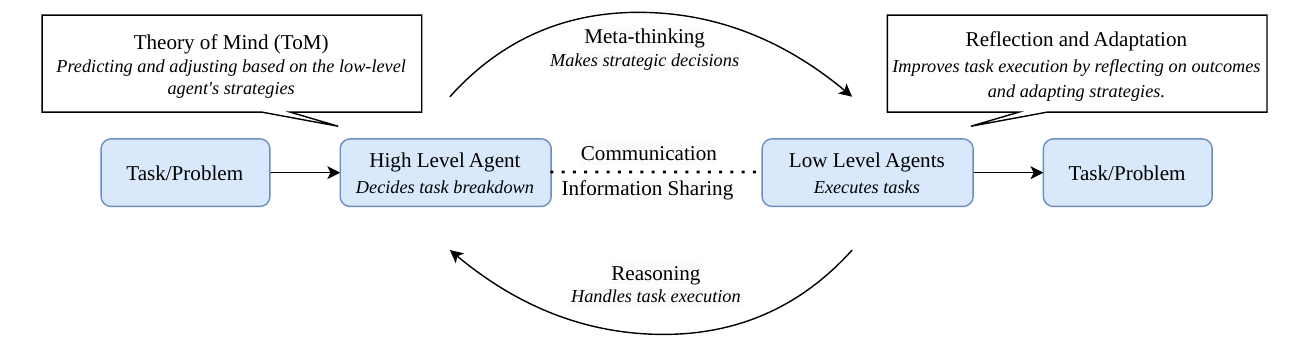} 
    \caption{The diagram illustrates a multi-agent system where a high-level agent breaks down tasks and communicates with low-level agents to execute them. The high-level agent predicts and adjusts strategies using ToM, while low-level agents provide feedback through task execution. Reflection and adaptation enable continuous improvement by refining strategies based on outcomes.}
    \label{fig:flow}
\end{figure*}
\subsection{Meta-Thinking Techniques using Multi-Agent Methods}
To address the limitations of single-agent feedback loops, researchers have turned to multi-agent architectures. One important architecture is the Supervisor-Agent Architecture, in which a high-level "supervisor" agent manages multiple lower-level agents, each being an expert at a particular form of reasoning or problem-solving \cite{leibo2017multi}. The hierarchical structure facilitates the division of work: lower agents provide proposals or intermediate reasoning steps, and a supervisor agent reorganizes, evaluates, and may override ill proposals. These hierarchical systems tend to produce more uniform and understandable outputs, observed in experimental research \cite{shu2024towards}. The overall structure of such a system is illustrated in Figure \ref{fig:flow}, where a high-level agent uses ToM to assign tasks and adapt strategies based on the feedback from low-level agents. The agents then execute the subtasks and report back results, enabling the system to be capable of looking back, learning, and adapting future decisions.\\
\indent Another multi-agent strategy is Agent-Based Debate and Self-Critique, inspired by the idea that competing perspectives can uncover hidden errors in reasoning \cite{irving2018ai}. When multiple specialized agents critique or defend one another, LLMs generate more robust solutions. For example, debate-based LLM systems encourage adversarial questioning, compelling each agent to justify or refine its positions \cite{ziegler2022adversarial}. An extension of this paradigm includes role-playing agent systems, such as AutoGPT \cite{dai2023ad}. In these systems, multiple instantiated personas (e.g., a “Planner” agent, an “Evaluator” agent, and a “Research” agent) interact within a shared environment to refine outputs iteratively. This interplay creates a meta-thinking layer, where the LLM-as-agents collectively reason, critique, and synthesize knowledge beyond what a single-agent system might achieve.
\subsection{Emerging RL-Based Methods}
Both from the single-agent and multi-agent frameworks, RLHF is a crucial method for guiding LLMs toward more reliable meta-thinking \cite{christiano2017deep, ouyang2022training}. RLHF is a process of refinement in which the LLM is trained not only on labeled content, but also on human preferences, most commonly through ranked responses, to make its output align with desired qualities such as truthfulness, coherence, and ability for self-correction \cite{kumar2024training}. Instead of relying on the signals generated by the model only, RLHF exploits the responses of human annotators in a bid to induce desired behaviors such as coherent explanation, honesty, or self-correcting behavior, and penalizes undesirable characteristics such as hallucinations or self-contradictions. By iterative tuning, LLMs are trained to incorporate both external human decisions and internal mechanisms of reflection, eventually increasing their ability for reflection and error identification. For example, OpenAI's ChatGPT-4 is trained using RLHF, where human evaluators label their responses as clear, precise, and aligned with ethical standards \cite{ouyang2022training}. When the model generates deceptive or biased content, human annotators provide corrective feedback, which instructs the model to better its subsequent output. Through a series of rounds of iterations, this enhances the model's ability to identify and eliminate inconsistencies to generate more believable responses.\\
\indent Another innovation is adaptive self-rewarding systems, which blend intrinsic motivations (e.g., consistency, curiosity) with extrinsic signals (e.g., supervisor-agent or human feedback). For instance, researchers \cite{beck2023survey} present a system where an LLM rewards itself with "meta rewards" for showing improved internal consistency or discovering new solution paths in a reasoning task. When used in multi-agent environments, where AI models collaborate or compete, the self-rewarding approach can help overcome common limitations of single-agent systems, which often get stuck in repetitive or suboptimal reasoning patterns. An AI model strengthens its reasoning skills and becomes more adaptable over time by continuously evaluating and refining its own reward system.\\
\indent These emerging methods underscore the ongoing shift from static, single-entity LLMs toward dynamic, interactive systems capable of sustained introspection. By integrating human feedback, hierarchical supervision, and intrinsic self-rewarding signals, researchers are progressively constructing AI systems that can more effectively critique, adapt, and refine their own reasoning processes.
\section{MARL Strategies for Meta-Thinking}
\begin{figure*}[h]
    \centering
    \includegraphics[width=\linewidth]{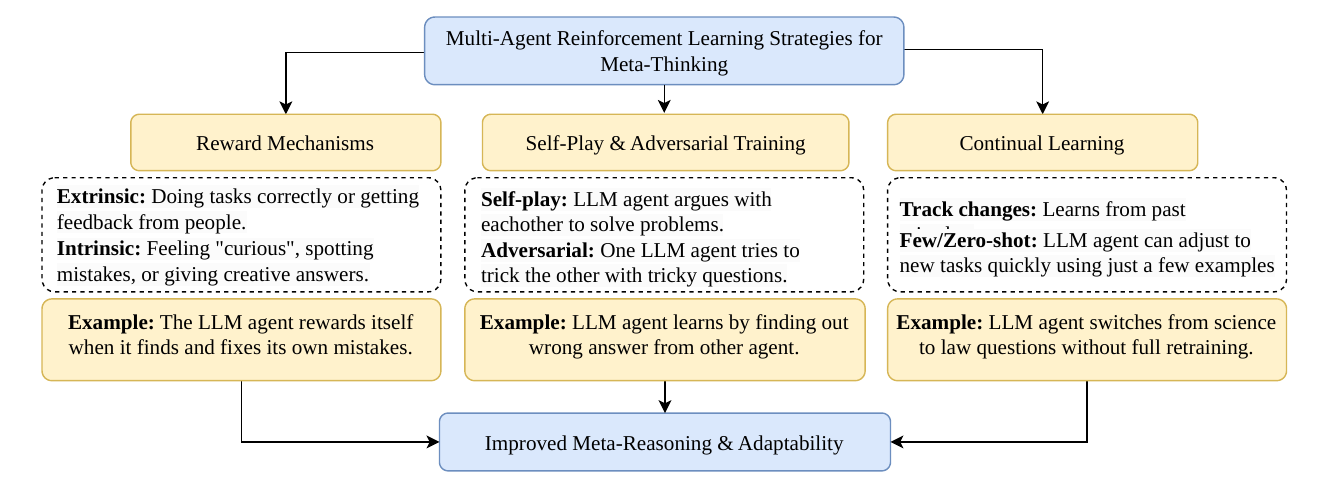} 
    \caption{Overview of RL Techniques Enabling Meta-Thinking in Language Models}
    \label{fig:techniques}
\end{figure*}

Meta-thinking in LLMs heavily relies on the architecture, presentation, and integration of reinforcement signals. Recent research suggests that well-structured reward mechanisms in RL, strategic self-play in MAS, and continuous meta-learning can significantly enhance a model’s ability to reflect on and improve its own reasoning and decision-making processes \cite{ma2023eureka, wu2024meta}. This section, as shown in Figure \ref{fig:techniques} describes three key strategies of RL solutions to meta-thinking: \\
\textbf{1) Designing Intrinsic and Extrinsic Meta-Rewards} i.e. for balancing human feedback with curiosity driven reward to drive self reflection, \\
\textbf{2) Collaborative Self-Play and Adversarial Training} i.e. agent debase, collaborate or challenge each other to uncover hidden flaws and deepen reasoning, and\\ 
\textbf{3) Meta-Learning for Continuous Adaptation} i.e. fast adaptation to new tasks by learning  ``how to learn'' across different domains.
\subsection{Designing Intrinsic and Extrinsic Meta-Rewards}
Designing rewards that encourage meta-thinking is about finding a balance between intrinsic and extrinsic incentives. Extrinsic rewards come from external sources, such as completing tasks or receiving human feedback, which traditional RL frameworks prioritize \cite{stiennon2020learning}. Intrinsic reward refers to a signal generated within the system as a substitute for external (extrinsic) reward from the world. It is generally based on quantities like novelty, prediction error, or gained information, and is designed to drive behavior like exploration, learning from curiosity, and policy optimization in scenarios where extrinsic feedback is sparsely timed, delayed, or unhelpful. Traditionally, RL frameworks for language modeling concentrate on extrinsic cues, such as achieving a task or receiving human feedback to guide the model updates \cite{stiennon2020learning}. Researchers increasingly believe that intrinsic motivation is the key to helping language models refine their reasoning \cite{li2024confidence}.  For instance, a model can reward itself for detecting and correcting contradictions in its thought process, fostering active error detection, self-diagnosis, and continuous learning.\\
\indent To make these reward systems effective, strict criteria are established to determine and evaluate the correctness, coherence, and novelty of the model’s responses \cite{ma2023eureka}. For example, a recent study by \cite{sun2025curiosity} introduces a novelty bonus, an intrinsic reward criterion that incentivizes the generation of creative or less conventional outputs. This approach helps discourage repetitive or stereotypical responses that lack depth. Collectively, such reward mechanisms not only improve task performance but also foster a more reflective reasoning style, where the model learns to evaluate and enhance its own thought process.
\indent As discussed in \cite{zheng2024online}, Intrinsic and extrinsic meta-rewards in LLMs involves combining external task signals with internal self-assessment signals into a single training objective. Concretely, at each step $t$, the model receives an \emph{extrinsic reward} $r^e_t$ (e.g., human preference or task completion) and an \emph{intrinsic reward} $r^i_t$ (e.g., novelty, coherence, or contradiction-detection), and forms the \emph{meta-reward}
\begin{equation}
  R_t = \lambda\,r^e_t + (1 - \lambda)\,r^i_t, \quad \lambda \in [0, 1],
\end{equation}
where:
   \( \lambda \in [0, 1] \) is a scalar balancing the two reward components.
For training stability, extrinsic rewards are normalized across the batch using the mean \( \mu \) and standard deviation \( \sigma \), following Group Relative Policy Optimization (GRPO)~\cite{shao2024deepseekmath}:
\begin{equation}
  \tilde{r}^e_i = \frac{r^e_i - \mu}{\sigma},
\end{equation}
where \( \mu \) and \( \sigma \) are the batch-wise mean and standard deviation of extrinsic rewards \( \{r^e_i\} \)~\cite{stiennon2020learning}. Intrinsic rewards can be generated by the language model itself. For instance, an ``LLM-as-judge''~\cite{chua2025learning} framework evaluates each generated token \( y_t \) in context \( y_{<t} \) using an internal judgment function \( J(\cdot) \):
\begin{equation}
  r^i_t = J\left(y_t \mid y_{<t}\right),
\end{equation}
with these token-level scores distilled into a lightweight reward model through asynchronous feedback mechanisms~\cite{li2024confidence}.
The policy \( \pi_\theta(a_t \mid s_t) \), parameterized by weights \( \theta \), defines the probability of selecting action \( a_t \) given the current state \( s_t \). In language models, \( s_t \) typically represents the current dialogue context or hidden state. The policy is optimized to maximize the expected discounted return:
\begin{equation}
  \theta^* = \arg\max_\theta \; \mathbb{E}_{\pi_\theta}\!\left[\sum_{t=0}^T \gamma^t\left(\lambda\,r^e_t + (1 - \lambda)\,r^i_t\right)\right],
\end{equation}
where: \( \gamma \in [0, 1] \) is the discount factor applied to future rewards, \( s_t \) is the current state (e.g., dialogue history or embedding context), \( a_t \) is the action taken (e.g., token generation), \( \pi_\theta \) is the policy distribution used by the model to choose actions. This formulation enables training language models that align with human objectives (\( r^e_t \)) while also leveraging self-generated critiques (\( r^i_t \)) to improve coherence, reasoning, and exploration.
\subsection{Collaborative Self-Play and Adversarial Training}
Multi-agent self-play has contributed significantly towards promoting emergent meta-think actions by focusing on the effectiveness of intrinsic and extrinsic reward schemes. In self-play setups, two or more replicas or implementations of an LLM either co-operatively or competitively interact with one another in a bid to solve sophisticated tasks \cite{vinyals2019grandmaster}. LLMs engage in multi-turn argumentation to enhance reasoning, math problem-solving to verify logical correctness, code writing and debugging from adversarial collaboration, strategic game playing to develop long-term planning, and scientific hypothesis testing to enhance structured reasoning. A good example is the study in \cite{meta2022human}, where the multi-agentic system achieved state-of-the-art performance in the complex strategy game Diplomacy by using a combination of LLMs and planning algorithms to negotiate, ally, and betray human players. The system used self-play training to reinforce its strategic communication and decision-making iteratively, demonstrating how interaction among multiple agents can give rise to emergent behavior that would not be observed in a single-agent setting. In this setup of multi-turn argumentation, LLMs learn to reason more strategically when they engage with each other, especially when encouraged to criticize or outperform each other. This process uncovers hidden strategies and develops advanced reasoning abilities. Therefore, models become better at anticipating challenges, identifying fallacies in arguments, and refining their responses based on counterexamples given by their opponents \cite{wu2024shall}. This give-and-take interaction enhances their power to reason, learn, and develop over time.\\
\indent A particular subfield of self-play is adversarial training, where adversarial agents generate test questions or scenarios designed to expose flaws in the primary model's logic \cite{cheng2024self}. These adversarial examples range from subtle inconsistencies in logic, such as asking a model to explain why "a square has three sides," to more complex contradictions, such as presenting contradictory historical timelines and asking the model to reconcile them. In mathematical reasoning, an adversarial agent might provide a subtly incorrect proof and ask the model to prove or disprove it. LLMs develop a more robust and self-skeptical form of reasoning—one that is better able to detect and revise subtle errors before LLMs have a chance to propagate by being iteratively exposed to these adversarial examples. This cycle of challenge and revision not only improves task performance but also reinforces meta-cognitive capacities by compelling the model to continually re-evaluate its own internal reasoning \cite{nakamura2023logicattack}.\\
\indent As in \cite{cheng2024self}, \textbf{Attacker} $\mathcal{A}_\theta$ (a function parameterized by $\theta$ representing the attacker policy) takes as input a hidden target word $w$ (the correct answer in the language game) and generates a prompt $x$ (a natural language cue used to guide the defender), such that:
\[
  x = \mathcal{A}_\theta(w)
\]
\textbf{Defender} $\mathcal{D}_\theta$ (sharing the same parameters $\theta$, acting as the defender agent) attempts to recover the original target word by interpreting the prompt to produce a predicted word $\hat{w}$:
\[
  \hat{w} = \mathcal{D}_\theta(x)
\]
A scalar reward function $R(w, \hat{w}) \in \mathbb{R}$ (which quantifies the success of the defender's inference) is defined based on prediction accuracy:
\[
  R(w, \hat{w}) = 
  \begin{cases}
    +1, & \text{if } \hat{w} = w \text{ (successful inference)} \\
    -1, & \text{otherwise}
  \end{cases}
\]
Model parameters $\theta$ are updated using reinforcement learning, such as Proximal Policy Optimization (PPO), to maximize the expected reward over a distribution of target words $\mathcal{W}$ (i.e., the task dataset), where $\alpha$ is the learning rate and $\nabla_\theta$ denotes the gradient with respect to parameters:
\[
  \theta \leftarrow \theta + \alpha \nabla_\theta \mathbb{E}_{w \sim \mathcal{W}} \left[ R(w, \mathcal{D}_\theta(\mathcal{A}_\theta(w))) \right]
\]
By this, self-play, LLMs refine both their prompt generation and inference strategies under adversarial conditions, which leads to improved reasoning and meta-cognitive performance \cite{cheng2024self}.
\subsection{Meta-Learning for Continuous Adaptation}
Above discussed reward mechanism and adversarial training can cause meta-thinking in the short run, for example, self-play-based RL has been shown to improve reasoning in mathematical problem-solving, where models iteratively refine their logic by engaging in adversarial debate, as discussed by \cite{wu2024meta}.  Meta-RL aims to enable fast adaptation to novel knowledge and reasoning patterns in the long run by helping models rapidly generalize to novel reasoning patterns and unseen knowledge. For example, \cite{wan2025rema} introduced a meta-learning framework where LLMs track past reasoning failures and dynamically refine their strategies and allow LLMs to adapt to new challenges, such as transitioning from scientific reasoning to legal document analysis, without retraining \cite{finn2017model}. In meta-learning, an LLM is trained to quickly optimize its inner parameters in response to new tasks or domains, in a way learning "how to learn." Such a shift is especially crucial for LLMs that must be versatile in ever-changing environments where data distributions or objectives may change abruptly. \indent The core idea of meta-RL is to create a feedback learning loop in which the model receives reinforcement not only from outcomes of tasks but also from meta-reward signals that evaluate improvements in reasoning efficiency, adaptability, and coherence across tasks. Such reinforcement can be shaped by indicators like reduced prediction error over time, faster convergence on new tasks, or successful reuse of past reasoning paths in new domains \cite{wu2024meta, yin2020meta}. Overall, meta-learning RL unlocks a lifelong learning paradigm for LLMs, where agents are not just trained to complete tasks, but are continuously evolving learners. This capability is essential for fostering deep, reflective, and general-purpose meta-thinking abilities in language models operating in dynamic, high-variance environments.
\indent Real-world applications of meta-learning in LLMs are often few-shot and zero-shot learning boosts that allow models to generalize from a limited number of examples to completely novel tasks \cite{yin2020meta}. By enhancing the meta-learner, researchers can enhance its ability to modify its reasoning process or reward signals from a few supervised instances. This has led to remarkable improvements in the trust and depth of reasoning of LLMs \cite{wu2024meta}. \\
\indent For example, a recent study \cite{li2024meta} introduced Micre (Meta In-Context Learning for Relation Extraction), which enhances LLMs' few-shot and zero-shot learning. In in-context learning, LLM generalizes on the few examples, meta learning principle teach the model to "learn how to learn" from the examples more effectively. Micre meta-train the LLMs to learn to adapt quickly with fewer examples. In Micre, LLM is not training just to perform tasks but also learning to adapt to new tasks. By reformulating relation extraction as a natural language generation task and meta-training over a broad variety of datasets, Micre does significantly better than traditional fine-tuning methods. As discussed in \cite{li2024meta}, In meta-RL, the model learns an adaptive policy $\pi_\theta$ parameterized by $\theta$, with the goal of fast adaptation to new tasks.\\
\textbf{Meta-training phase:} The model is trained over a distribution of tasks $\mathcal{T} \sim p(\mathcal{T})$. For each task $\mathcal{T}_i$, the model performs inner-loop adaptation using a small dataset $\mathcal{D}^{\text{train}}_i$:
\[
\theta_i' = \theta - \alpha \nabla_\theta \mathcal{L}_{\mathcal{T}_i}^{\text{train}}(\theta)
\]
where $\theta_i'$ are the task-adapted parameters, and $\alpha$ is the inner-loop learning rate.
\textbf{Meta-objective:} Update the original parameters $\theta$ based on performance over a validation set $\mathcal{D}^{\text{test}}_i$:
\[
\theta \leftarrow \theta - \beta \nabla_\theta \sum_{\mathcal{T}_i \sim p(\mathcal{T})} \mathcal{L}_{\mathcal{T}_i}^{\text{test}}(\theta_i')
\]
where $\beta$ is the meta (outer-loop) learning rate. At inference time, given a small support set $\mathcal{S}$ (few-shot examples), the model produces a prediction $\hat{y}$ conditioned on in-context demonstrations:
\[
\hat{y} = \mathcal{M}_{\theta}(x \mid \mathcal{S})
\]
Here, $\mathcal{M}_\theta$ learns to perform task adaptation internally by interpreting $\mathcal{S}$ during inference.
This formulation enables the LLM to generalize to new reasoning patterns and domains with minimal data and no parameter updates.
\section{Evaluation Metrics, Comparative studies and Datasets}
Assessing meta-reasoning in LLMs is key to developing systems that reason, self-correct, and adapt like humans. To achieve this, researchers have developed targeted metrics that challenge models' ability to introspect and revise their own reasoning. This section introduces key evaluation metrics for assessing meta-reasoning in LLMs, followed by a summary of experimental results from studies comparing MARL in LLMs. Lastly, it presents specialized datasets designed to support and enhance meta-reasoning evaluation as shown in Table \ref{tab:meta_reasoning_summary}.
\subsection{Metrics for Evaluating Meta-Thinking in LLMs}
Many metrics have been proposed to assess LLM’s ability for meta-thinking, including 1) logical consistency, 2) self-correction ability, and 3) reasoning depth. \\
\indent \textbf{Logical‑consistency} metrics \cite{raj2025improving,cobbe2021training} measure whether a model’s multi‑step reasoning stays contradiction‑free, for example, correctly deriving $6kg$ for $2kg × 3kg$ and not later claiming the total is 8kg.\\
\indent \textbf{Self-correction ability} is a characteristic of models equipped with robust meta-thinking, enabling them to detect and rectify their own errors. Self-correction can be measured by injecting small perturbations or contradictions into prompts and observing if the model identifies the discrepancy and corrects itself \cite{wen2024red}. This approach not only tests error detection but also probes the model’s ability to reflect on its output and make improvements. \\
\indent \textbf{Reasoning depth} assesses the degree to which a model is able to breakdown complex problems into systematic sub steps. For example, \cite{wei2022chain} recommends that one adds a COT score to determine whether a model's intermediate reasoning steps provide sufficient detail and coherence. High depth of reasoning could indicate high meta-thinking, as it requires persistent monitoring and self-correction of one's train of thought.\\
\indent \textbf{ELA} \cite{zeng2024mr} measures how precisely an LLM can pinpoint mistakes in a candidate COT across 5,975 process-based questions. For each question, the LLM is shown an automatically generated CoT solution and asked to 1) classify it as correct/incorrect, 2) identify the step where the first error occurs, and 3) explain that particular error. The results in \cite{zeng2024mr} shows that State-of-the-art (e.g. OpenAI’s o1 series \cite{jaech2024openai}) achieve over 70\% ELA, while many open-source LLMs remain below 45\%, revealing a substantial gap in fine-grained error detection.  \\
\indent \textbf{Meta- vs. Object-Level Metrics} \cite{ferguson2025evaluating} separate high-level planning from low-level inference: 1) Meta-Level Reasoning Frequency is the fraction of cases with explicit planning steps. 2) Object-Level Reasoning Accuracy is correctness on the detailed inference sub-steps. The results is \cite{ferguson2025evaluating} show that LLMs exhibit greater than 85\% Meta-Level Frequency but drop to 60–70\% on Object-Level Accuracy, indicating they can plan well but often face difficulty in the low-level inferences.\\ 
\indent \textbf{Depth-Wise Accuracy} \cite{patel2024multi} tracks performance drop as logical inference depth increases. The increase in the depth means that there is an increase in the number of chained inferences. For example, Depth 3 means “If A then B; if B then C; if C then D; given A and C, is D true?” The findings in \cite{patel2024multi} show that when Depth is 1, LLMs have an approximate average accuracy of 68\%, but then when Depth becomes 5, the accuracy falls to 43\% approximately. The paper \cite{patel2024multi} noted the trend that an LLM often fails at integrating conflicting premises.\\   
\indent \textbf{AIA} and \textbf{MIA} \cite{sonkar2024malalgoqa} quantify an LLM’s ability to choose the correct rationale and to spot flawed (“malgorithm”) reasoning paths \cite{sonkar2024malalgoqa}. 1) AIA is the percentage of cases where the model picks the true rationale for the correct answer. 2) MIA is the percentage of cases where the model identifies the flawed rationale behind a wrong choice. The findings in the paper \cite{sonkar2024malalgoqa} LLMs achieve approximately 75\% AIA but drop to approximately 45\% MIA and show that LLMs struggle more with recognizing faulty reasoning. \\  
\indent \textbf{SEQ} \cite{hada2024metal} measures the quality of the generated summary by the LLM model. It encompasses: 1) Linguistic Acceptability: To measure grammar accuracy and fluency. 2) Task Quality: Ability to accomplish the summarization task. 3) Output Content Quality: Content relevance and accuracy. 4) Hallucinations: Absence of constructed or ungrounded information. 5) Problematic Content: Avoid offensive, biased, or unsafe language.
\subsection{Comparative Studies of MARL-Enhanced LLMs}
A growing body of comparative studies highlights the advantages of incorporating MARL into LLMs for improved meta-thinking:\\
\indent \textbf{Hierarchical MARL for Meta-Reasoning:} \cite{wan2025rema} introduce a multi-agent configuration where a "supervisor" agent coordinates multiple "worker" agents, each with partial sub-problems of a reasoning problem. Experiments on the StrategyQA \cite{geva2021did} tasks demonstrate enhanced coherence and fewer contradictions compared to single-agent baselines.\\
\indent \textbf{Agent-Based Debate and Self-Critique:} Experiments of \cite{zhang2024can} examine adversarial debates between LLM "defender" and "prosecutor" agents. They note a dramatic enhancement of logical consistency scores—derived from the DebateQA dataset—following models' exposure to several adversarial interactions.\\
\indent \textbf{Coordination vs. Competition:} \cite{leibo2021scalable} conduct ablation experiments to compare cooperative MARL settings with purely competitive ones. They find that cooperative approaches offer better measures of self-reflection (e.g., higher internal consistency), with competitive setups being superior at detecting subtle flaws. Periodic switches between cooperative and adversarial phases in hybrid strategies tend to yield the best trade-off, fostering both robust error detection and collaborative improvement.

\begin{table*}[ht]
  \centering
  \caption{Summary of Evaluation Metrics, Comparative Studies, and Datasets for Meta-Reasoning in LLMs}
  \label{tab:meta_reasoning_summary}
  \begin{tabularx}{\textwidth}{@{}l l X l@{}}
    \toprule
    \textbf{Category}        & \textbf{Name}                                & \textbf{Description}                                                                                  & \textbf{References}                          \\
    \midrule
    \multirow{8}{*}{Metric}  & Logical Consistency & Detects internal contradictions in a model's COT, ensuring step-by-step coherence. & \cite{raj2025improving}  \cite{cobbe2021training}   \\
    \cmidrule(lr){2-4}
                             & Self-correction Ability                      & Measures whether a model spots and fixes injected errors or contradictions in its own output.         & \cite{wen2024red}                            \\
    \cmidrule(lr){2-4}
                             & Reasoning Depth                              & Assesses granularity and completeness of intermediate reasoning steps in multi-step problems.         & \cite{wei2022chain}                          \\
    \cmidrule(lr){2-4}
                             & Error Localization Accuracy                  & How precisely an LLM pinpoints mistakes in reasoning chains (MR-Ben).                                  & \cite{zeng2024mr}                         \\
    \cmidrule(lr){2-4}
                             & Meta-/Object-Level Metrics                   & Planning frequency vs. inference accuracy (Franklin Dataset).                                        & \cite{ferguson2025evaluating}                  \\
    \cmidrule(lr){2-4}
                             & Depth-Wise Accuracy                          & Performance drop as logical inference depth increases (Multi-LogiEval).                               & \cite{patel2024multi}                \\
    \cmidrule(lr){2-4}
                             & AIA / MIA                                    & Correct rationale vs. flawed reasoning identification (MalAlgoQA).                                   & \cite{sonkar2024malalgoqa}                      \\
    \cmidrule(lr){2-4}
                             & Correlation Scores                           & Agreement between LLM evaluators and humans in multilingual summarization (METAL).                    & \cite{hada2024metal}                         \\
    \midrule
    \multirow{3}{*}{Comparative Study} 
                             & Hierarchical MARL for Meta-Reasoning         & Supervisor-worker hierarchies decompose tasks, improving coherence on StrategyQA.                     & \cite{wan2025rema,geva2021did}               \\
    \cmidrule(lr){2-4}
                             & Agent-Based Debate \& Self-Critique           & Defender vs. prosecutor debates boost logical consistency on DebateQA.                               & \cite{zhang2024can}                          \\
    \cmidrule(lr){2-4}
                             & Coordination vs. Competition (Hybrid)       & Alternating cooperative/adversarial phases balance reflection and flaw detection.                     & \cite{leibo2021scalable}                     \\
    \midrule
    \multirow{9}{*}{Dataset} 
                             & BIG-Bench                                    & 204 tasks probing reasoning, self-reflection, bias detection, and perspective-shifting.               & \cite{srivastava2022beyond}                  \\
    \cmidrule(lr){2-4}
                             & SciInstruct                                  & STEM problems with built-in self-critique steps for scientific reasoning.                             & \cite{zhang2024sciinstruct}                  \\
    \cmidrule(lr){2-4}
                             & DebateQA                                     & approximately 3,000 debate-style questions to test balanced argumentation.                                          & \cite{xu2024debateqa}                        \\
    \cmidrule(lr){2-4}
                             & StrategyQA                                   & 2,780 implicit multi-step reasoning questions with annotated inference chains.                        & \cite{geva2021did}                           \\
    \cmidrule(lr){2-4}
                             & MR-Ben                                       & 5,975 process-based questions measuring error localization in chains of thought.                      & \cite{zeng2024mr}                         \\
    \cmidrule(lr){2-4}
                             & Franklin Dataset                             & QA reframed into meta- vs. object-level tasks; tracks planning vs. inference.                       & \cite{ferguson2025evaluating}                  \\
    \cmidrule(lr){2-4}
                             & Multi-LogiEval                               & Logical reasoning across 30+ rules at depths 1–5; reports depth-wise accuracy.                        & \cite{patel2024multi}                \\
    \cmidrule(lr){2-4}
                             & MalAlgoQA                                    & Counterfactual QA with metrics for correct vs. flawed reasoning paths (AIA/MIA).                     & \cite{sonkar2024malalgoqa}                      \\
    \cmidrule(lr){2-4}
                             & METAL                                        & Multilingual, reference-free summarization evaluation; reports correlation scores vs.\ human judges.  & \cite{hada2024metal}                         \\
    \bottomrule
  \end{tabularx}
\end{table*}
\begin{figure}[h]
    \centering
    \includegraphics[width=0.75\linewidth]{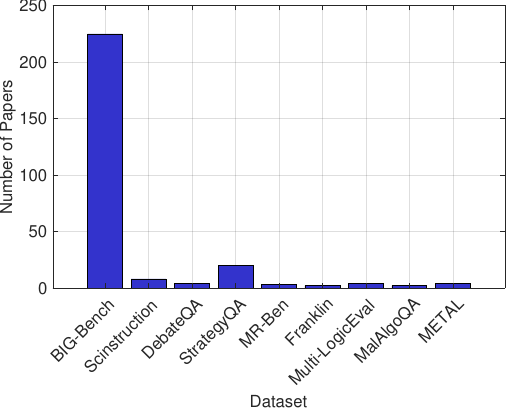} 
    \caption{Number of published papers referencing each dataset for evaluating LLM meta-reasoning.}
    \label{fig:rl_papers}
\end{figure}
\subsection{Existing Datasets}
Over the past few years, researchers have introduced specialized datasets and benchmarks to evaluate self-reflection and iterative reasoning in LLMs. Figure \ref{fig:rl_papers} illustrates the relative popularity of each dataset, measured by the number of published papers incorporating them in the experiment. Notably, BIG-Bench \cite{srivastava2022beyond} appears in more studies compared to other datasets, reflecting its broad applicability in testing a wide range of language-model capabilities. In contrast, specialized datasets such as SciInstruct \cite{zhang2024sciinstruct} and MR-Ben \cite{zeng2024mr} have fewer overall citations but target deeper skills such as domain-specific scientific reasoning or process-based meta-cognition. DebateQA \cite{xu2024debateqa}, StrategyQA \cite{geva2021did}, FRANKLIN \cite{ferguson2025evaluating}, Multi-LogiEval \cite{patel2024multi}, MalAlgoQA \cite{sonkar2024malalgoqa}, and METAL \cite{hada2024metal} similarly occupy more focused niches within the broader landscape of LLM evaluation. The brief introduction of the dataset is given below. \\
\indent \textbf{BIG-Bench} \cite{srivastava2022beyond} dataset is designed to ascertain the ability of language models to reason, think, and even question their own outputs. The dataset is comprised of 204 different challenges, ranging from solving mathematics and languages to recognizing bias and even creating stories. The interesting part of the dataset is that some of the tasks force LLMs to reflect on their thinking, like, "Are you sure about your answer?" or "Can you find a flaw in what you've stated?" For example, in one of the tasks, LLM systems are asked to justify a joke, a difficult task because humor involves world knowledge and reasoning. Another challenge involves the LLM responding to a question, and then rewriting its response from a different perspective. \\
\indent \textbf{SciInstruct} \cite{zhang2024sciinstruct} is a dedicated data set to enhance the scientific reasoning ability of language models. It is based on subjects like physics, chemistry, mathematics, and formal proofs to enable the LLMs to resolve higher level scientific challenges. Self-reflection is one of the fundamental features of SciInstruct where LLMs learn to inspect and improve their own reasoning mechanisms. Through this self-criticism, the LLM becomes more accurate and reliable. By incorporating self-critical practices, SciInstruct aims to develop LLMs that can better understand and solve complex scientific problems.\\
\indent \textbf{DebateQA} \cite{xu2024debateqa, mohsin2025retrieval} is a set created to test the extent to which AI can answer tough questions in which nobody has one "right" answer—just like real debates. It has nearly 3,000 questions with different human-sourced answers reflecting different sides of the issue. For example, the question "Is climate change mainly caused by human activities? " has some yes answers that point to causes like pollution and deforestation, and others that respond with no, citing events like volcanic eruptions or solar changes. The idea is to check if LLM understands that questions like these are contentious, and if it can present multiple viewpoints instead of one.".\\
\indent \textbf{StrategyQA} \cite{geva2021did} is a dataset used to test the ability of LLMs to respond to questions that involve implicit, multi-step reasoning. In contrast to simple questions, those in StrategyQA tend to require the LLMs to bridge several pieces of information to reach an answer. For example, the query "Did Aristotle use a laptop?" requires understanding that Aristotle lived centuries ago and laptops are a modern invention, leading to the answer "No." The dataset consists of 2,780 of these kinds of questions, each with a related breakdown of the steps of reasoning and supporting evidence from Wikipedia.\\
MR‑Ben \cite{zeng2024mr} is a 5,975-question benchmark for meta‑reasoning: models must spot, explain, and fix errors in step‑by‑step solutions (e.g., swapping $x/y$ vs $y/x$ in Graham’s‑law effusion). Covering logic, chemistry, programming, and more, it tests deliberate “System‑2” thinking; results show GPT‑4‑o excels at self‑correction while most other models struggle, underscoring the need for stronger meta‑cognitive skills.\\
\indent \textbf{FRANKLIN Dataset} \cite{ferguson2025evaluating} access how LLMs handle two types of reasoning: meta-level (planning and strategizing) and object-level (actually doing tasks like calculations). For instance, a question such as "Which European country had the highest GDP growth in 2023?" requires the model to devise a plan of action first (meta-level) and then fetch specific GDP data and compare (object-level). The study found that while LLMs are proficient at declaring strategies, they fall short when it comes to executing specific tasks, usually getting data fetching or calculations incorrect. \\
\indent \textbf{Multi-LogiEval} \cite{patel2024multi} spans 30+ logical inference rules report Depth-Wise Accuracy. The Multi-LogiEval dataset introduces a dataset to assess LLMs to determine whether LLMs can address complex, multi-step logical reasoning. The dataset includes problems across three logic types: 1) propositional logic (simple if-then statements), 2) first-order logic (objects and relations between them), and 3) non-monotonic logic (where conclusions can change on the addition of new information). For example, a question would be: "If all birds fly, and penguins are birds, do penguins fly?" In which the model must use rules of logic to reach the correct conclusion. The study found that the LLMs' accuracy is inversely proportional to the number of reasoning steps, showing challenges on the part of the LLMs to perform deep logical reasoning.\\
\indent \textbf{MalAlgoQA} \cite{sonkar2024malalgoqa} is a QA benchmark built to probe multi‑step reasoning. It contains 7 logic types (e.g., deductive, abductive, numerical, counterfactual) and poses multi‑context questions that require chaining facts e.g., “All machines are robots and RoboX is a machine, Is RoboX a robot?”, with AIA/MIA metrics distinguishing sound vs. flawed reasoning.\\
\indent \textbf{METAL} \cite{hada2024metal} offers 1000 GPT‑4–generated summaries (100 each in 10 languages) for reference‑free quality evaluation. Native speakers and LLMs (GPT‑3.5‑Turbo, GPT‑4, PaLM 2) rate grammar, fluency, and task relevance, enabling cross‑lingual correlation studies.
\section{Challenges and Open Problems}
\subsection{Challenges}
Despite remarkable advances in the training of meta-thinkers with MARL, there are challenges. The challenges cut across computational constraints, system stability, and ethics, each with seminal research questions awaiting novel solutions.
\subsubsection{Scalability and Stability in MARL for Meta-Thinking} 
One of the central challenges with MARL-based meta-thinking is the computational burden of dealing with many agents. As each agent contributes to the entire reasoning framework, the framework is likely to become too heavy to manage successfully \cite{du2021survey}. For instance, in hierarchical MARL frameworks, interactions between complex-task supervisors and low-level worker agents grow in number exponentially \cite{wan2025rema}. Therefore, training such systems requires substantial memory and processing resources, rendering these systems impractical for several real-world uses.\\
\indent In addition, coordination complexity is caused when multiple agents interact concurrently. Agents need to negotiate shared policies, exchange partial solutions, or provide criticism. This requires extensive message-passing protocols that, unless efficiently optimized, result in bottlenecks and synchronization issues. These scalability issues are usually solved by using techniques such as distributed training, hierarchical architecture, or inter-agent communication compression \cite{papoudakis2019dealing}. But how much such approaches compromise between computational complexity and robust meta-thinking remains questionable.\\
\indent MARL systems are also susceptible to mode collapse, a phenomenon where agents converge prematurely to suboptimal solutions, often because homogeneous strategies control the learning process \cite{mohebbi2023games}. This is especially problematic for meta-thinking tasks, for which diversity of perspective and self-criticism are essential to unearth hidden flaws. If all agents learn the identical flawed patterns of reasoning, then the potential for meta-thinking is severely diminished.\\
\indent Another such challenge is reward hacking, in which agents find loopholes to earn large rewards without actually improving reasoning \cite{denison2024sycophancy}. On a meta-thinking scale, reward hacking can be considered inflation of self-correcting behavior agents can continue to add meaningless mistakes and "correct" them to accomolate self-improvement rewards. Therefore, Stable training dynamics necessitate well-constructed reward functions that reflect improved meta-cognitive abilities. Researchers continue to study novel penalty functions, adaptive reward schedules, and curriculum learning methods to discourage exploitative strategies \cite{yang2021diverse}.
\subsubsection{Energy Efficiency and Resource-Aware Design in MAS} 
MAS, particularly RL-driven MAS, are likely to experience high energy consumption challenges. Communication between the agents to synchronize is one of the main factors, which is particularly costly for large or highly dynamic environments \cite{ahmad2002multi}. RL agents, with computational demands, dynamic environment interaction, and policy updating requirements, add to the extra energy consumption \cite{10.1007/978-3-031-41734-4_26}.\\
\indent Also, the majority of MAS remain to apply time-triggered data collection, where LLM agents sense and transmit data at intervals regardless of changes in the environment or the need for urgency in tasks. This leads to unnecessary data transmission and energy loss \cite{dorri2018multi}. Task allocation is a second issue. Task allocation protocols tend to ignore the processing power or energy available to agents at a moment and therefore overload low-resource agents and waste energy on the system \cite{krothapalli2002distributed}.
 \subsubsection{Ethical and Safety Considerations}
As MARL-driven LLMs become more proficient at self-improvement, ethics and safety become paramount. Agents that can learn their own decision-making can also learn to perpetuate bias or disseminate hateful material if their reward signals or agent interactions reinforce unwanted patterns \cite{weidinger2022taxonomy}. For example, if the supervisory agent of a model is derived from a biased dataset, it can propagate biased reward signals across the entire multi-agent system, which can then be compounded with existing prejudices.\\
\indent Additionally, giving unbiased self-improvement consistently is a non-trivial problem. When human critics provide real-time feedback, the danger of unconscious bias—cultural, gender, or racial—envelops the model's objectives. Hence, designing open audit trails, high-quality fairness measures, and human-in-the-loop systems that detect and compensate for biases are research areas in active pursuit \cite{mitchell2019model}. The ultimate aim is to reconcile the freedom and efficiency of MARL systems with demands for safety, responsibility, and inclusivity in AI implementation.
\subsection{Open Problems}
Based on the above issues, research is increasingly looking to interdisciplinary knowledge and hybrid solutions to augment meta-thinking skills. The next decade will witness progress on multiple fronts, from neuroscience-driven frameworks to more interpretable models that combine MARL with symbolic reasoning.
\subsubsection{Neuroscience-Inspired Approaches for Meta-Thinking}
One optimistic avenue is inspiration from human cognition systems where meta-cognition occurs due to emergent interactions across numerous brain regions \cite{hassabis2017neuroscience}. They have begun thinking about architectures mimicking human mechanisms of learning, such as in the form of episodic memory modules or gated information by attention—so the introspection capacity of an LLM can be enhanced \cite{bengio2017consciousness}. Through the application of neuroscientific knowledge to agent construction, designers expect to enable more human-like self-awareness and flexible response to novel tasks. Such biologically motivated systems would be especially beneficial in high-stakes situations requiring robust problem-solving combined with awareness of one's own mental activity.
\subsubsection{Hybrid Models that Combine MARL and Symbolic Reasoning}
Another significant direction focuses on enhancing interpretability in AI reasoning by means of hybrid models integrating MARL and symbolic logic \cite{acharya2023neurosymbolic}. Symbolic modules can embody domain-specific rules or knowledge graphs, while MARL agents are responsible for discovering and applying these rules in a more adaptive manner. This integration has several advantages: symbolic reasoning can help to reduce reward hacking risk by applying logical constraints, and MARL can handle uncertainty or partial information best over strictly symbolic systems.\\
\indent For example, a multi-agent system can utilize symbolic constraints to verify consistency in each reasoning step, while RL agents cope with broader exploration of solution approaches \cite{bhuyan2024neuro}. Early testing indicates that these hybrid methods can also significantly improve transparency because human judges can trace back how a single symbolic inference led to better decision-making or self-correction \cite{yang2018peorl}. Optimizing such hybrid designs and demonstrating their success in a broad range of tasks is a fine research area in the coming years.
\subsubsection{Scaling Multi-Agent Architectures for General AI}
Finally, as multi-agent paradigms prove themselves in niche domains, researchers are moving away from LLMs to general multi-agent AI systems. The aim is to build adaptive agents capable of generating, arguing, and proving solutions for real-world problems such as robotics, finance, and healthcare \cite{ross2023programmer, bilal2025onrl}. These domains require more than linguistic proficiency; they require robust situational awareness, sensor fusion, and real-time decision-making under uncertainty.\\
\indent Achieving General AI through multi-agent interaction involves increasing the number of agents, specialising agents for different tasks, and developing emergent "societies" of agents that can dynamically self-adapt \cite{leibo2021scalable}. While the computational complexities are high, the reward potential is AI systems that not only learn solutions but also maintain improving collective problem-solving abilities in the face of changing problems. This wide-ranging vision of multi-agent cooperation and self-improvement heralds an exciting frontier in the search for advanced meta-thinking in artificial intelligence.
\section{Conclusion}
In this survey, we have systematically examined the emergence of meta‑thinking in LLMs through the lens of MARL. We began by diagnosing the key limitations of existing self‑reflection approaches such as, hallucinations, unchecked error propagation, and domain specificity, and then surveyed both single‑agent techniques (e.g., COT prompting, self‑distillation, RLHF) and multi‑agent architectures (supervisor‑worker hierarchies, adversarial debates, self‑play) that can endow LLMs with robust introspective capabilities. Building on these foundations, we detailed essential MARL strategies—designing intrinsic and extrinsic meta‑rewards, collaborative self‑play and adversarial training, and meta‑learning for continual adaptation that collectively allow LLMs to evaluate, refine, and adapt their own reasoning in real time. We also reviewed the state of evaluation metrics and datasets for meta‑reasoning, and identified persistent challenges around scalability, stability, energy efficiency, and ethical safety. Finally, we outlined a future roadmap encompassing neuroscience inspired architectures, hybrid symbolic MARL systems, and scalable multi‑agent “societies,” all aimed at realizing trustworthy, self‑correcting AI. We hope this comprehensive framework will guide future research toward LLMs that not only perform complex tasks but also continually think about and improve their own thought processes.
\bibliographystyle{IEEEtran}
\bibliography{main,Bibliography}  
\end{document}